**CohortFinder: an open-source tool for data-driven partitioning of biomedical image cohorts to yield robust machine learning models.**

Fan Fan[1], Georgia Martinez[2], Thomas Desilvio[2], John Shin[2], Yijiang Chen[2], Bangchen Wang[3], Takaya Ozeki[4], Maxime W. Lafarge[5], Viktor H. Koelzer[5], Laura Barisoni[3,6], Anant Madabhushi[1,7], Satish E. Viswanath[2], Andrew Janowczyk[1,8,9]

[1]Emory University and Georgia Institute of Technology, Department of Biomedical Engineering, Atlanta, Georgia, USA
[2]Case Western Reserve University, Department of Biomedical Engineering, Cleveland, Ohio, USA
[3]Duke University, Department of Pathology, Division of AI & Computational Pathology, Durham, North Carolina, USA
[4]University of Michigan, Department of Internal Medicine, Division of Nephrology, Ann Arbor, Michigan, USA
[5]University Hospital of Zurich, University of Zurich, Department of Pathology and Molecular Pathology, Zurich, Switzerland
[6]Duke University, Department of Medicine, Division of Nephrology, Durham, North Carolina, USA
[7]Atlanta Veterans Administration Medical Center, Atlanta,
Georgia, USA
[8]University Hospital of Geneva, Department of Oncology, Division of Precision Oncology, Geneva, Switzerland
[9]University Hospital of Geneva, Department of Clinical Pathology, Division of Clinical Pathology, Geneva, Switzerland

**Abstract**

Batch effects (BEs) refer to systematic technical differences in data collection unrelated to biological variations whose noise is shown to negatively impact machine learning (ML) model generalizability. Here we release CohortFinder[1], an open-source tool aimed at mitigating BEs via data-driven cohort partitioning. We demonstrate CohortFinder improves ML model performance in downstream medical image processing tasks. CohortFinder is freely available for download at cohortfinder.com.

**Main**

The increased availability of digital pathology (DP) whole slide images (WSI) and radiographic imaging datasets has propelled the development of both machine and deep learning algorithms to aid in disease diagnosis, patient prognosis, and predicting therapy response [1]. These algorithms work by identifying patterns in digital data that are associated with clinical outcomes of interest. While large-scale data analysis was previously limited by storage, processing, and computational constraints, modern-day development and test of these models increasingly involves the collection of large cohorts over both physical (e.g., institutions) and temporal (e.g., time points) spaces [1]. However, differences in non-biological preanalytical processes at these various spatiotemporal points likely impart undesirable batch effects (BE) in the final digital data. For example, BEs in DP images generated in the same manner from the same tissue type yield significant visual differences which may impact data interpretation (shown in **Figure 1-A**).

In DP, these BEs tend to originate from, but are not limited to, differences in physical processes (tissue processing, storage, glass slide preparation) as well as digitization processes (scanners, color profile management, compression approaches) [1~6]. In MR imaging cohorts, these BEs may result from MRI acquisition protocols, patient preparation differences, or imaging artifacts such as noise, motion, inhomogeneity, ringing, or aliasing [7,8]. Regardless of the modality, BEs have been shown to not only severely hamper both the peak performance and robustness (i.e., degree of performance change when examining new unseen data) of machine learning (ML) models but can also cause spurious discoveries when associated with outcome variables of interest[5,9,10] For example, ML has been shown to be able to detect and thus be influenced by BEs associated with site of origin [12], potentially leading to biased accuracy in prediction of features including survival, genomic mutations, and tumor stage. Given the detrimental impact of BEs, there have been several approaches aimed at ameliorating them. For example, stain/color normalization/augmentation approaches were developed to decrease color-based BEs of WSI [2-6,11]. Similarly, in radiographic imaging, image standardization and feature normalization approaches (e.g., ComBat) have been proposed to reduce BE-related variability [7,8,13,14]. However, the inconsistent performance and wide variability of approaches utilized suggests that a more systematic accounting of BEs during the cohort partitioning phase of ML model construction may be warranted. Partitioning in this context is defined as the ML best practice of dividing a cohort into training and testing sets. The training set is employed to create the model, while the testing set is used to determine the model's generalizability performance on previously unseen data not employed during training.

---

[1] The link to CohortFinder: http://cohortfinder.com

By far the most common way to partition cohorts is by randomly assigning patients to training and testing sets, which we term the **Average Case (AC).** The AC strategy, however, has the potential to expose its employer to *unreasonably* sub-optimal cohort partitions by sheer misfortune. For example, in a **Worst Case (WC)**, images demonstrating similar BEs may be mutually exclusively assigned to the *same* training/testing split, resulting in the training data sharing minimal or no visual similarity with the testing set (see **Extended Data Figure 1-(3)**). Such a cohort partition is likely to result in maximally exposing its user to the deleterious effects of BEs, and thus yielding significantly inferior performance of the resulting ML model on the testing set. Although this WC is an unlikely event, it represents the end point of a continuum of potential real-world sub-optimal ACs, wherein models are exposed to only a subset of the true range of BE variability in the data. It then stands to reason that there is likely a **Best Case (BC)** on the opposite end of the continuum which maximally balances BEs to yield more representative data partitions in turn yielding more generalizable ML models.

Towards addressing BEs in biomedical imaging data, we have developed and released CohortFinder, an open-source data-driven partitioning tool for specifically determining BC cohort partitions for training and testing ML models. CohortFinder consumes quality control (QC) metrics (e.g., via HistoQC/MRQy [8,11], two open-source QC tools for DP and MR imaging data), and at the patient level, performs unsupervised clustering to determine **BE-groups** which are strikingly homogenous in presentation (see **Extended Data Figure 1-(2)).** By iteratively partitioning these BE groups at a user-defined ratio into training and testing sets, CohortFinder yields highly representative and diverse partitions, which balance BEs, even in cases of minority BE groups. CohortFinder also provides the ability, when given relevant spatiotemporal labels (e.g., site origin, date of scan) or downstream outcome labels (e.g., good/poor prognosis), to statistically test for BEs and provide an associated report. To evaluate the ability of CohortFinder to yield BC data partitions, three different deep-learning use cases in DP and radiographic imaging are evaluated here: (a) tubule segmentation on kidney WSIs, (b) adenocarcinoma detection on colon WSIs, and (c) rectal cancer segmentation on MR images (see **Extended Data Table 1**).

For quantitative comparisons, five commonly used evaluation measures (Precision, Recall, Accuracy, IoU, and F1-score), were calculated to compare the performance of BC, AC, and WC partitioning via internal cross-validation as well as on external testing data (i.e., 1 patient from each different site or scanner) for all three use-cases separately. In **Extended Data Table 2 & Extended Data Table 3**, the overall performance (average and standard deviation) and fold-specific values for each evaluation measure are reported, respectively. From the tables, WC partitioning demonstrates the worst quantitative performance in all evaluation measures compared to AC and BC partitioning, across all use cases. For example, for the colon adenocarcinoma classification use case**,** BC demonstrates an average F1-score improvement of *0.23* compared to WC in the external testing dataset. Further, BC also results in a relatively lower standard deviation than AC for most evaluation measures, suggesting that CohortFinder can aid in producing more robust ML models which exhibit less variance. The violin plot shows the distribution of the F1-score discussed above (**Figure 2**). From the violin plot, all the metrics of WCs are more dispersed as compared to the AC and BC. In most cases, while the distribution of BC is often more compact than AC, occasionally the distribution between AC and BC is similar. This observation supports the notion that data partitions generated via random sampling (i.e., AC) exist on a spectrum of BE mitigation, with some providing better or worse accounting. This spectrum also demonstrates that a user employing random sampling has no way of knowing where their partitions lie on the BE mitigation spectrum and may in fact be experiencing a WC partition by chance. In contrast, CohortFinder provides users with the assurance that they have an idealized BE accounting while requiring only minutes of computational overhead. **Figure 2** further enables qualitative comparisons of ML model results between partitioning scenarios (WC/AC/BC) for all three use cases. For the tubule segmentation & colon adenocarcinoma classification tasks, there are fewer FPs and FNs in the BC results than those in AC and WC. For the rectal tumor segmentation use case, BC best predicts the tumor contour compared to AC and WC.

Together, these results suggest that CohortFinder provides a systematic partitioning strategy that yields ML models with improved performance and generalizability. Given the low computational burden associated with its usage (i.e.,1-2 minutes on a consumer-grade laptop), we believe it will serve as a valuable tool for the avoidance of employing suboptimal WC-like cohort partitions, which may result from the typically used approach of random sampling when creating data partitions.

To summarize, we have presented and released an open-source data-partitioning tool termed CohortFinder. CohortFinder works by identifying potential batch-effect groups and ensuring their proportional representation when partitioning a cohort into training and testing sets, yielding demonstrably more reliable downstream ML models in batch-effect-laden datasets. Importantly, CohortFinder ingests input metrics in a common CSV format, produced by open-source quality control tools (HistoQC/MRQy [8,11]). This suggests that as our knowledge of batch effects and quality control improve, and more sophisticated metrics are developed, CohortFinder will be organically capable of leveraging them for further improving downstream ML models. CohortFinder is released to the public for further community review, comment, and usage at cohortfinder.com.

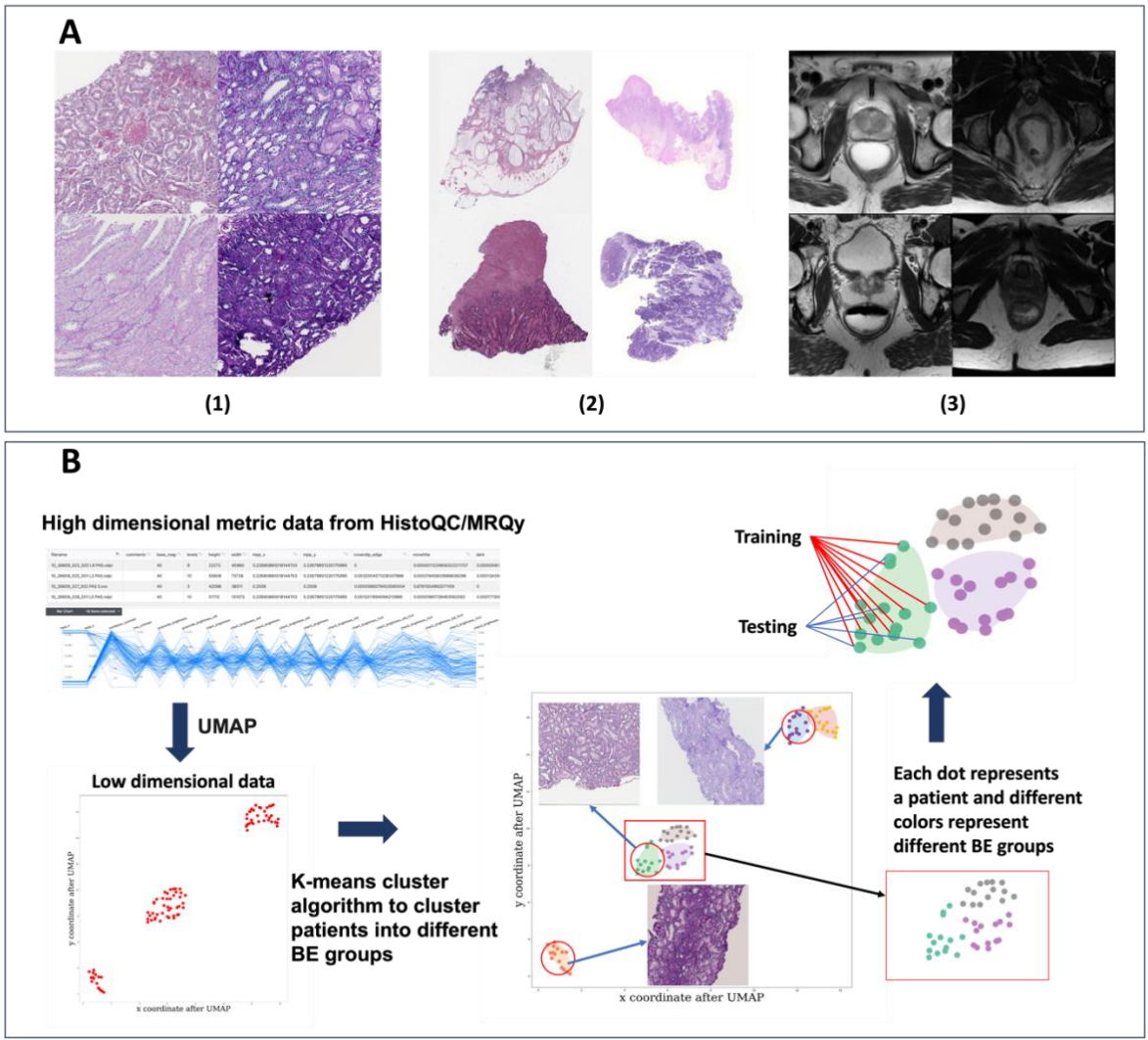

Figure 1 **A:** Examples of the batch effects with 1) four ROIs from the tubule segmentation task, 2) four WSI thumbnails from the colon adenocarcinoma detection task, 3) four slice images of four different patients from the rectal cancer segmentation task. As can be seen, the DP images show notable differences in white balance, brightness, and contrast demonstrating clear BEs. Similarly, the MR imaging data also shows significant differences in foreground contrast. **B:** The basic workflow for CohortFinder. First, UMAP is used to project high-dimensional quality control metric values into a 2-dimensional space. Second, k-means clustering takes place in this 2-dimensional space to identify BE-groups using approximately k target clusters. Finally, patients in each BE group are assigned into training/testing set based on the user-given ratio while sampling from each BE group.

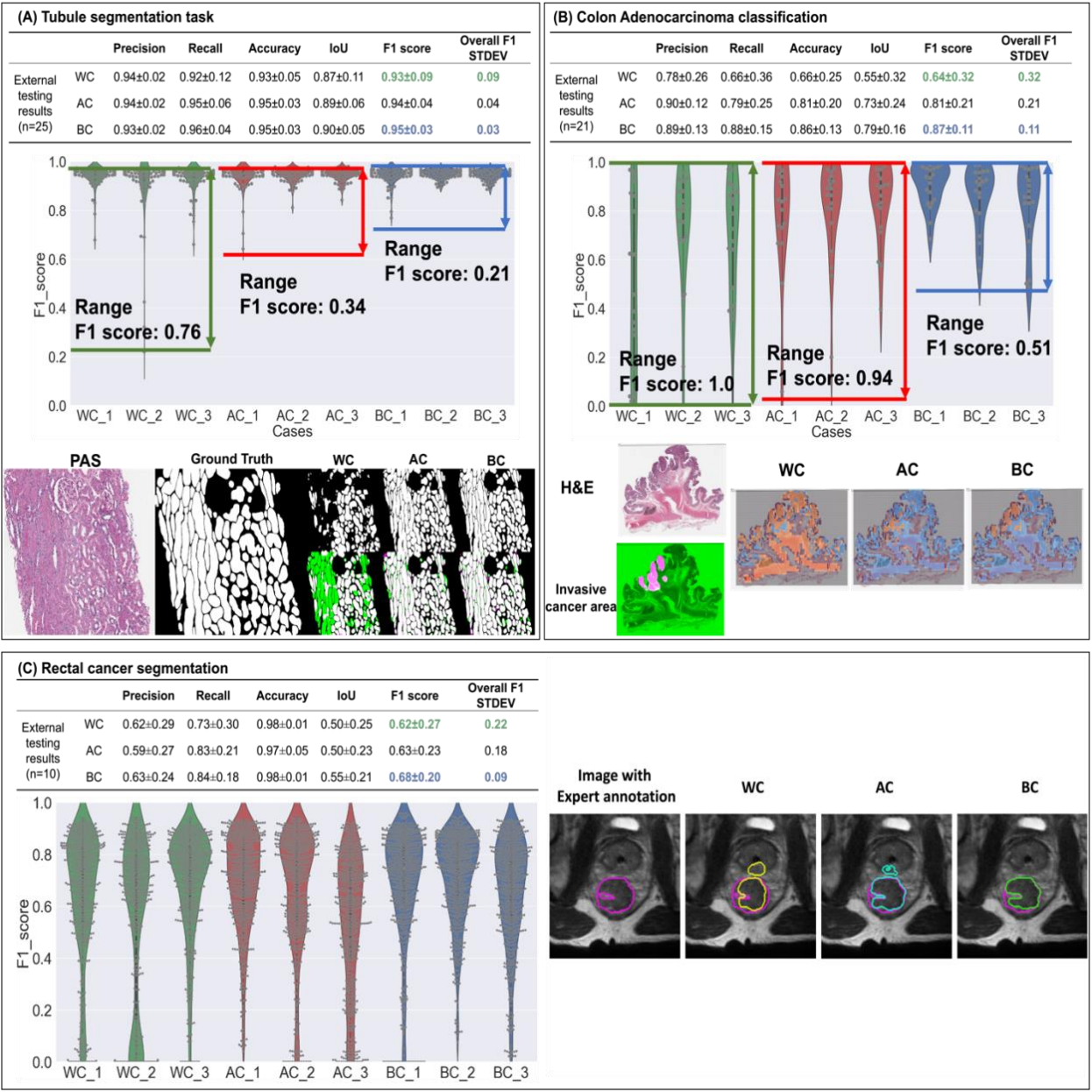

Figure 2 **Quantitative and qualitative results for all three use cases**. For each use case, we have **(1)** The overall performance on internal and external testing datasets. **(2)** F1 score performance for 9 different models on the external testing datasets, where the gray dots in each violin plot indicate individual performance for a single image. **(3)** Qualitative results. In **tubule segmentation task (A)**, the first column is a cropped PAS-stained image, the second column is the tubule segmentation ground truth (GT), and the remaining images are the results of WC, AC, and BC. In each scenario, the top row is the DL model result, while the bottom row corresponds to the overlay image between DL output images and the GT, where green parts represent the false negative (FN) area, and the red parts represent the false positive (FP) area. WC has more FN and FP areas compared to AC & BC. Compared to AC, BC has fewer FN and FP areas. For **colon cancer classification task (B)**, the images in the first column are the H&E thumbnails and cancer annotation (the tumor area in pink, non-tumor in green). The remaining three images are the heatmaps for the WC/AC/BC, where the orange area represents the predicted cancer area and blue represents the predicted no-cancer area, and the gray area represents the non-informative area (background/non-tissue area). From the heatmaps, WC underpredicted the tumor regions, AC overpredicted the tumor region, while BC yields the best overlap between tumor area and ground truth. **Rectal cancer segmentation task (C)**, the first column is the image with expert annotation ground truth in pink, which is also shown as a pink contour in the remaining three columns. The 2D U-net segmentation results for WC (yellow), AC (cyan), and BC (green) show that WC and AC overpredicted the tumor region while BC marginally underpredicted. **In all three tasks**, violin plots of F1 scores show a decreasing trend from BC to AC to WC. AC is also seen to have a larger F1 score range, lower average F1 value, and a higher standard deviation than BC; suggesting AC performance is less robust than BC.

**Code availability**
CohortFinder is available at: https://cohortfinder.com

**Acknowledgements**
Research reported in this publication was supported by the NCI under award numbers U01CA239055, 1R01LM013864, 1U01DK133090, U01CA248226, R01DK118431, 1U01CA248226, the National Institute of Nursing Research (1R01NR019585-01A1), the NIBIB through the CWRU Interdisciplinary Biomedical Imaging Training Program Fellowship (2T32EB007509-16), the DOD Peer Reviewed Cancer Research Program (W81XWH-21-1-0345, W81XWH 19-1-0668, W81XWH-21-1-0725), the Wen Ko APT Summer Internship Program, the Ohio Third Frontier Technology Validation Fund, and the Wallace H. Coulter Foundation Program in the Department of Biomedical Engineering at Case Western Reserve University and sponsored research funding from Pfizer. This work made use of the High-Performance Computing Resource in the Core Facility for Advanced Research Computing at Case Western Reserve University.

The Nephrotic Syndrome Study Network (NEPTUNE) is part of the Rare Diseases Clinical Research Network (RDCRN), which is funded by the National Institutes of Health (NIH) and led by the National Center for Advancing Translational Sciences (NCATS) through its Division of Rare Diseases Research Innovation (DRDRI). NEPTUNE is funded under grant number U54DK083912 as a collaboration between NCATS and the National Institute of Diabetes and Digestive and Kidney Diseases (NIDDK). Additional funding and/or programmatic support is provided by the University of Michigan, NephCure Kidney International and the Halpin Foundation. RDCRN consortia are supported by the RDCRN Data Management and Coordinating Center (DMCC), funded by NCATS and the National Institute of Neurological Disorders and Stroke (NINDS) under U2CTR002818. The results of the histology analysis of colorectal cancers shown here utilize digital pathology images generated by the TCGA Research Network: https://www.cancer.gov/tcga.

Research reported in this publication relating to Dr Madabhushi's contributuin was supported by the National Cancer Institute under award numbers R01CA268287A1, U01CA269181, R01CA26820701A1, R01CA249992-01A1, R01CA202752-01A1, R01CA208236-01A1, R01CA216579-01A1, R01CA220581-01A1, R01CA257612-01A1, 1U01CA239055-01, 1U01CA248226-01, 1U54CA254566-01, National Heart, Lung and Blood Institute 1R01HL15127701A1, R01HL15807101A1, National Institute of Biomedical Imaging and Bioengineering 1R43EB028736-01, VA Merit Review Award IBX004121A from the United States Department of Veterans Affairs Biomedical Laboratory Research and Development Service the Office of the Assistant Secretary of Defense for Health Affairs, through the Breast Cancer Research Program (W81XWH-19-1-0668), the Prostate Cancer Research Program (W81XWH-20-1-0851), the Lung Cancer Research Program (W81XWH-18-1-0440, W81XWH-20-1-0595), the Peer Reviewed Cancer Research Program (W81XWH-18-1-0404, W81XWH-21-1-0345, W81XWH-21-1-0160), the Kidney Precision Medicine Project (KPMP) Glue Grant and sponsored research agreements from Bristol Myers-Squibb, Boehringer-Ingelheim, Eli-Lilly and Astrazeneca.

**Conflict of interest statement**

AM is an equity holder in Picture Health, Elucid Bioimaging, and Inspirata Inc. Currently he serves on the advisory board of Picture Health, Aiforia Inc, and SimBioSys. He also currently consults for SimBioSys. He also has sponsored research agreements with AstraZeneca, Boehringer-Ingelheim, Eli-Lilly and Bristol Myers-Squibb. His technology has been licensed to Picture Health and Elucid Bioimaging. He is also involved in 3 different R01 grants with Inspirata Inc. VHK has acted as an invited speaker for Sharing Progress in Cancer Care (SPCC) and Indica Labs, is on an advisory board of Takeda, and holds sponsored research agreements with Roche and IAG unrelated to the content of this study. SV has received sponsored research funding from Pfizer. AJ provides consulting for Merck, Lunaphore, and Roche, the latter of which he also has a sponsored research agreement. LB is a consultant for Sangamo and Protalix and is on the scientific advisory boards of Vertex and Nephcure.

**Contributing Author Statement**

F.F. contributed to the study conceptualization and design, data curation, interpretation of data, experiment execution and drafting of the manuscript; G.M., T.D., J.S, Y.C., B.W., T.O., M.W.L, V.H.K and S.E.V contributed to the study design, data curation, interpretation of data, and experiment execution; A.J., L.B., S.E.V, and A.M. contributed to the methodology, code base, study conceptualization and design, gathering of resources, data curation, interpretation of data, reviewing, and editing.

**Supplementary Materials**

**Methods**

**Data partitioning based on BE groups**

CF proceeds by identifying multiple BE "groups", i.e., sets of images with similar presentation metrics as calculated by HistoQC/MRQy[8,11] (two open-source quality control tools for pathology and radiology, respectively). These BE groups are then iteratively randomly divided into subsets at user-specified ratios (the ratio of the testing data and all the data). As a result, the training and testing sets have balanced representations of presentation profiles to help to ensure diversity for ML models.

A key component in consistently generating BC data partitions is the ability to detect BEs *a priori*. While this can be approximated using either available metadata (e.g., site or scanner labels) or by visual assessment, the labor and labeling involved quickly becomes intractable and non-reproducible for large datasets. This also does not leverage the most critical source of information readily available, the presentation of the images themselves. Importantly, previous work has demonstrated that computationally derived quality control (QC) metrics can be repurposed to detect BEs [8~11,13].

**HistoQC/ MRQy functionality**

CohortFinder utilizes the output from either HistoQC or MRQy, open-source tools designed to aid in QC of DP and MRI images, respectively [8,11]. These tools allow for large-scale high-throughput extraction of deterministic image quality measures. In both modalities, images are sequentially fed into the pipeline where each module (a) captures basic metadata (e.g., the base magnification, the microns per pixel, repetition time, echo time, number of slices per volume), (b) quantifies visual characteristic metrics (e.g., brightness, contrast, mean of the foreground, contrast per pixel), and (c) locates artifacts (e.g., air bubbles, pen markings, noise, and inhomogeneity). The resulting metrics form the input for CohortFinder and are used to identify BE groups.

**CohortFinder functionality**

The basic CohortFinder workflow (see **Figure 1-B**), proceeds as follows:

(a) CohortFinder loads the extracted QC measures.

(b) The QC measures are considered a high-dimensional vector and projected into 2 dimensions via UMAP [16] for visualization at a 2-dimensional embedded plot. UMAP works by modeling the data as a fuzzy topological manifold, which allows it to capture complex relationships between data points that might be missed by linear techniques like PCA. It then optimizes a low-dimensional projection of the data, aiming to preserve both the local and global structure of the manifold. UMAP was selected as it can incorporate new data points into an existing embedding without recomputing from scratch.

(c) *k*-means clustering takes place in this 2-dimensional space to identify *k* target clusters where each cluster is considered to represent a BE group, replicated clustering was used here to mitigate the impact of k-means' randomness and improve the stability of the clustering algorithm.

(d) for each BE group, the set of images assigned to it is randomly assigned to the training and testing sets at a ratio provided by the user.

CF produces four outputs: (1) UMAP plots (shown in **Extended Data Figure 2**) indicate colored BE-group distribution results in 2-dimensional UMAP space based on QC measures (**Extended Data Figure 2-a**). (2) Patient assignment results for training and testing (**Extended Data Figure 2-b**) where "v" indicates a patient to be placed in the training set versus "o" to indicate the testing set, (2) a contact sheet type image (shown in **Extended Data Figure 3**) with representative images from each BE-group, (3) a general log containing information for the user as well as potential errors, and (4) a comma separated value file which contains (a) the metrics used to perform the BE-group detection, (b) the resulting UMAP coordinates, (c) the determined BE-group index number, and (d) the label assigned to the particular image (e.g., training vs testing).

**Batch-effect testing module of CohortFinder**
If the user also provides labels of interest (e.g., known site of origin, or clinical variable of interest), CohortFinder will run statistical t-tests, similar to previous works [17], to determine if there are selected feature(s) (QC metrics which are used to define BEs) that are statistically significantly confounding the label. Additionally, a random forest (RF) model is employed as part of a permutation test [10] to estimate BE prediction ability from input metrics. These metrics are then ranked by the RF results to help users assess the importance of each feature in detecting BEs.

**Experimental design**
To evaluate the ability of CohortFinder to yield optimal data partitions, three different deep-learning use cases in DP and medical imaging areas were selected: (a) tubule segmentation on kidney WSIs, (b) adenocarcinoma detection on colon WSIs, and (c) rectal cancer segmentation on MR images (see **Extended Data Table 1**). For each use case, 1 patient from each site/scanner was randomly selected to be included in a benchmark external testing set while the remaining patients were used for developing training and testing partitions. For the latter, three scenarios were explored:

- Best Case (BC): Patients were segregated via CohortFinder into BE groups, following which equitable BE distribution of samples was ensured in all partitions. The number of clusters (k) in CohortFinder was set to the number of patients divided by 3, to allow for 3-fold cross-validation where 1 patient from each

3-patient cluster is assigned to 1 of the 3 cross-validation folds (BC_1, BC_2, and BC_3). For example, in the tubule use case which had 91 patients, the CohortFinder-cluster parameter was set to 31.
- Average Case (AC): As per typical ML practices, samples were randomly split into 3 average-case folds (AC_1, AC_2, and AC_3) without considering BEs.
- Worst Case (WC): Patients were segregated to intentionally maximize the BE differences, as defined by HistoQC/MRQy metrics, between each worst-case group (WC_1, WC_2, and WC_3). To do so, patients were clustered into 3 BE groups (visually or by using CF), yielding groups that are notably different in presentation. Similar to how we determined BC, CohortFinder was used here to cluster the patients into 3 BE groups, where each BE group had a similar patient number. This allowed for 3-fold cross-validation where the patients in one BE group were assigned to 1 of the 3 cross-validation folds (WC_1, WC_2, and WC_3). For example, in the tubule use case which had 91 patients, each BE group was comprised of 31/30/30 patients.

During the experimental evaluation, internal patient-level cross-validation took place with each of the folds serving as the training set, and the derived model subsequently evaluated on the remaining folds as an "internal" testing set. For example, AC_1 was used to train an ML model which was then evaluated on AC_2 and AC_3. This internal testing process was conducted to gain a robust estimate of ML model performance within the training set. Patient-level distribution ensured that images from the same patient only appeared in a specific fold and were not distributed across folds. Each trained ML model was also tested on the external testing cohort, allowing for a fair cross-fold comparison.

**Evaluation metrics**

Five metrics were used to evaluate model performance: precision, recall, accuracy, IOU, and F1 score, based on their wide usage in ML model-based segmentation and classification tasks. Before calculating the metric value, true positive (TP), true negative (TN), false positive (FP) and false negative (FN) predictions were calculated at pixel level (for two segmentation use cases) and patch level (for the classification use case). Each metric was subsequently measured following the formulas in **Extended Data Table 2**.

**Network configuration and training**

U-net [18] was used for the segmentation tasks and Dense-Net [19] was used for the classification task, selected based on their popularity for these tasks. Both architectures were implemented in PyTorch with the following configuration details:

1. Tubule segmentation: **(a)** depth of the U-Net: 5 blocks, number of filters in the filter layer: 4, **(b)** patch size:512x512, number of training batches for each epoch: 6, **(c)** number of training epochs: 50, the model with the lowest validation loss was used to do the testing, **(d)** optimization algorithm: Adam and **(e)** data augmentation: vertical & horizontal flips and rotation were used during the network training process.

2. Colon adenocarcinoma classification: **(a)** Dense-Net architecture: growth rate is 32, drop rate is 0, initial feature number is 64, batch norm size is 2, **(b)** patch size: 224, number of training batches for each epoch: 64, **(c)** number of training epochs: 50, the model with lowest validation loss was used for the testing, **(d)** optimization algorithm: Adam and **(e)** data augmentation: vertical & horizontal flips and rotation were used during the network training process.
3. Rectal cancer segmentation: **(a)** depth of U-Net: 5 blocks, number of filters: 30, **(b)** size of cropped input images: 128 by 128, batch size: 16, **(c)** 50 epochs specified but with early stopping implemented based on the dice similarity coefficient loss function with a patience of 4 epochs, model with lowest validation loss was used for testing **(d)** optimization algorithm: Adam and **(e)** data augmentation: vertical & horizontal flips and rotation.

**Datasets**

### Tubule segmentation on kidney WSIs

#### *Cohort description*

Kidney WSIs were obtained from the Nephrotic Syndrome Study Network (NEPTUNE) digital pathology repository. NEPTUNE is a multisite observational cohort study of children and adults with glomerular disease, enrolled at the time of a clinically indicated kidney biopsy [17,22,25]. The renal biopsies were processed in 38 different pathology laboratories, collected, and shipped to the NEPTUNE image coordinating center, where glass slides were centrally scanned by two scanners (Aperio Scanscope AT2, Leica Biosystems Inc., Buffalo Groove, IL, USA and Hamamatsu Nanozoomer 2.0 HT, Hamamatsu Corporation, Hamamatsu City, Japan; both with an Olympus UPlan-SApo 20X objective, with a 0.75 NA, and image doubler) and subsequently uploaded into the NEPTUNE DPR [25,26]. In this use case, 116 Periodic acid–Schiff (PAS) stained WSIs from 25 sites were included. WSIs were chosen such that each patient contributed one randomly selected WSI.

#### *Ground-truth generation*

For each WSI, 3 ROIs (region of interest) from the cortical area containing tubule primitive were randomly selected and manually cropped as 3000 x 3000 tiles at 40x digital magnification. A pre-trained DL model [27] was implemented on each tile to acquire a rough tubule segmentation result. Two renal pathologists then manually evaluated all tubule segmentation results and revised them until all the tiles' results reached at least less than 5% FN/FP at tubule level. **Extended Data Figure 4** is one example of the tubule segmentation use case.

### Colon adenocarcinoma classification on colon WSIs

#### *Cohort description*

Colon WSIs were obtained from The Cancer Genome Atlas (TCGA) Colon Adenocarcinoma (COAD) cohort, a publicly available repository (https://portal.gdc.cancer.gov/projects/TCGA-COAD). 26 WSIs were excluded during the quality control (QC) process because of low base magnification or bad quality. For this use case, a total of 352 whole slide images (WSIs) were selected, all of which contained diagnostic information indicating the presence of 'Adenocarcinoma, NOS'. WSIs were chosen such that each patient contributed one WSI.

### *Ground-truth generation*

For every WSI, a senior pathologist manually annotated representative areas of colorectal adenocarcinoma to be used in an image-based molecular classification task [33], and the training ground-truth patches were generated based on this expert annotation. For the patch sampling process, HistoQC [11] was used to generate the tissue mask image (**Extended Data Figure 5–A. (1)**) for each WSI, with all the potential patches (256 x 256 at 5x magnification, no overlap) being generated from it before being further assigned to certain ground truth label by tessellation (**Extended Data Figure 5–A. (2)**). Patches were retained for training if (a) >90% area was intersecting with the detected tissue mask, and (b) were non-white (i.e., background removal by color thresholding). Patches were labeled as positive for cancer if >90% of the patch fell within the ground-truth annotation. **Extended Data Figure 5-B** shows representative cancer and non-cancer patches from one WSI, which were utilized for experimental evaluation.

## Rectal cancer segmentation on MRIs

### *Cohort description*

MRI scans were acquired from 166 patients diagnosed with rectal adenocarcinoma between August 2007 and October 2015, who had been retrospectively accrued from two institutions (University Hospitals Cleveland Medical Center and Cleveland Clinic, OH, USA). Based on IRB approval, informed consent was waived as all data had been deidentified prior to analysis. These MRI scans had been acquired prior to neoadjuvant chemoradiation treatment for primary staging of the rectal tumor, via a T2-weighted turbo spin echo sequence (T2w) in the axial plane on ten unique scanners from two different manufacturers (Philips and Siemens). Despite scanner variability, imaging parameters were fairly consistent within each institution (in-plane resolution: 0.313-1.172mm, slice thickness: 3.0-8.0mm, repetition time: 2400-11800msec, echo time: 64-184msec).

### *Ground-truth generation*

Annotations of rectal tumor extent on each T2w MRI dataset were obtained from a radiologist at each institution, who had access to clinical, pathologic, and radiology reports, as well as any additional imaging planes and sequences. Radiologists annotated the entirety of the gross tumor volume on all 2D slices between the peritoneal reflection and the top of the anal canal using 3D Slicer [30]. To minimize the effect of resolution differences within this cohort, all patient datasets and corresponding tumor annotations were resampled to a common resolution of 1.00 x 1.00 x 1.00mm. After resampling, 2D slices without tumor annotations were excluded, while the remaining 7897 2D slices were cropped to a uniform 128x128 bounding box centered on the tumor region.

# Results

## Tubule segmentation

### *Quantitative result*

In the external testing set (n=25 patients), the best F1 overall score results were from BC partitioning (0.95±0.03), followed by AC (0.94±0.04), and finally WC (0.93±0.09), with statistically significant differences between WC and AC ($p<0.01$) as well as between WC and BC ($p<0.01$). While no statistically significant differences were observed between AC and BC (p=0.71), AC resulted in a larger range of F1 scores (the violin plot in **Figure 2-A**), a lower average F1 value, and a higher standard deviation in F1 scores (the table in **Figure 2-A**) compared to BC. This suggests less robust performance for AC compared to BC.

### *Qualitative result.*

In **Figure 2-A** and **Extended Data Figure 6-A**, WC partitioning results in a relatively higher number of false negative (FN) areas (overlayed green regions) in comparison to AC and BC. Additionally, AC yielded a marginally higher number of false positive (FP) and FN regions (highlighted in purple and green, respectively) when compared to BC.

## Colon adenocarcinoma classification

### *Quantitative result*

In the external testing dataset (n=21 patients), the F1 score is seen to be significantly lower between BC and WC (0.87±0.11 vs 0.64±0.32, $p<0.01$) as well as between AC and WC (0.81±0.21 vs 0.64±0.32, $p<0.01$). Though no significant differences were found between BC and AC (0.87±0.11 vs 0.81±0.21, p = 0.09), the violin plots in **Figure 2-B** and **Extended Data Figure 7** suggest that BC has a more compact F1 score distribution, a higher average F1 score, as well as a lower standard deviation compared to AC.

### *Qualitative result.*

In **Figure 2-B** and **Extended Data Figure 6-B**, classification heatmaps produced via BC partitioning exhibit the highest degree of similarity with the ground-truth mask. WC partitioning resulted in a significant underprediction of the tumor area, with a considerable number of false negative (FN) patches within normal tissue. AC partitioning yielded a slightly smaller prediction of tumor area when compared to BC.

## Rectal cancer segmentation

### *Quantitative result*

In the external testing dataset (n=10 patients), BC models resulted in the highest overall F1 score of 0.68±0.20, while the AC and WC models yielded significantly lower overall F1 scores of 0.63±0.23 ($p<0.01$ vs BC) and 0.62 ±0.20 ($p<0.01$ vs BC), respectively (**Figure 2-C**). The markedly higher standard deviation in F1 scores of the WC models is illustrated in the violin plots of **Extended Data Figure 7-C**. Notably, the bottom tails of the F1 score distribution for WC models (green) are seen to be longer and wider in comparison to those of the AC (red) and BC models (blue). This indicates that tumor segmentations by WC models were more varied and shared little overlap with expert annotations, resulting in these marked variations in model performance compared to AC and BC.

*Qualitative result.*

**Figure 2-C** and **Extended Data Figure 6-C** depict representative tumor segmentations obtained via WC, AC, and BC partitioning schemes compared to radiologist annotations for two different patients. BC tumor segmentations are seen to consistently overlap with expert delineations, while AC models appear to slightly over-segment the tumor region. By comparison, the WC model is seen to have more varied performance in terms of under-segmenting or over-segmenting the tumor.

# Extended data - figures and tables

| Task | Modality | Dataset | Description | Evaluation metric |
|---|---|---|---|---|
| Tubule segmentation | Digital Pathology PAS-stained WSI | NEPTUNE | N=116 WSIs were selected, originating from 25 different institutions. Each WSI represents one single patient. From each site, 1 patient is randomly chosen for the external testing dataset. | Precision Recall Accuracy IOU F1 score |
| Colon adenocarcinoma classification | Digital Pathology H&E-stained WSI | TCGA-COAD | N = 356 cases were selected diagnosed as 'Adenocarcinoma, NOS', originating from 21 different institutions. From each site, 1 patient is randomly chosen for the external testing dataset. | |
| Rectal Cancer segmentation | Radiographic Imaging MR image | University Hospitals and Cleveland Clinic | N = 166 patients' MRIs were accrued from University Hospitals Cleveland Medical Center and Cleveland Clinic, originating from 10 different MRI scanner machines. 10 patients (1 per MRI scanner) were chosen for the external testing dataset. | |

Extended Data Table 1 List of use cases and associated experiments employed for the evaluation of CohortFinder.

| Metric | Formula |
|---|---|
| Precision | $\frac{tp}{tp + fp}$ |
| Recall | $\frac{tp}{tp + fn}$ |
| Accuracy | $\frac{tp + tn}{tp + tn + fp + fn}$ |
| IOU | $\frac{tp}{tp + fn + fp}$ |
| F1 score | $\frac{2 \times Recall \times Precision}{Recall + Precision}$ |

Extended Data Table 2 The calculation formulas of the five quantitative evaluation metrics

|  |  |  | Precision | Recall | Accuracy | IoU | F1 score |
|---|---|---|---|---|---|---|---|
| Tubule segmentation | Internal testing results | WC | 0.93±0.04 | 0.91±0.14 | 0.93±0.06 | 0.85±0.13 | **0.91±0.10** |
|  |  | AC | 0.93±0.04 | 0.94±0.08 | 0.94±0.03 | 0.88±0.08 | **0.93±0.05** |
|  |  | BC | 0.92±0.05 | 0.95±0.06 | 0.94±0.03 | 0.88±0.07 | **0.93±0.05** |
|  | External testing results | WC | 0.94±0.02 | 0.92±0.12 | 0.93±0.05 | 0.87±0.11 | 0.93±0.09 |
|  |  | AC | 0.94±0.02 | 0.95±0.06 | 0.95±0.03 | 0.89±0.06 | 0.94±0.04 |
|  |  | BC | 0.93±0.02 | 0.96±0.04 | 0.95±0.03 | 0.90±0.05 | **0.95±0.03** |
| Colon adenocarcinoma classification | Internal testing results | WC | 0.66±0.38 | 0.56±0.40 | 0.56±0.32 | 0.46±0.36 | **0.54±0.38** |
|  |  | AC | 0.89±0.18 | 0.85±0.23 | 0.84±0.18 | 0.77±0.24 | 0.84±0.20 |
|  |  | BC | 0.88±0.18 | 0.88±0.21 | 0.85±0.16 | 0.79±0.23 | **0.86±0.19** |
|  | External testing results | WC | 0.78±0.26 | 0.66±0.36 | 0.66±0.25 | 0.55±0.32 | **0.64±0.32** |
|  |  | AC | 0.90±0.12 | 0.79±0.25 | 0.81±0.2 | 0.73±0.24 | 0.81±0.21 |
|  |  | BC | 0.89±0.13 | 0.88±0.15 | 0.86±0.13 | 0.79±0.16 | **0.87±0.11** |
| Rectal cancer segmentation | Internal testing results | WC | 0.69±0.26 | 0.66±0.28 | 0.97±0.05 | 0.48±0.23 | 0.62±0.24 |
|  |  | AC | 0.61±0.29 | 0.74±0.27 | 0.95±0.06 | 0.46±0.23 | **0.60±0.24** |
|  |  | BC | 0.67±0.25 | 0.75±0.26 | 0.96±0.05 | 0.51±0.22 | **0.64±0.22** |
|  | External testing results | WC | 0.62±0.29 | 0.73±0.30 | 0.98±0.01 | 0.50±0.25 | **0.62±0.27** |
|  |  | AC | 0.59±0.27 | 0.83±0.21 | 0.97±0.05 | 0.50±0.23 | 0.63±0.23 |
|  |  | BC | 0.63±0.24 | 0.84±0.18 | 0.98±0.01 | 0.55±0.21 | **0.68±0.20** |

Extended Data Table 3 Summary of performance measures for 3 different use cases, reported on both internal and external testing datasets. For the F1 score, the best performance is highlighted in red, and the worst performance is highlighted in blue.

| Tubule segmentation | | | | | | | | |
|---|---|---|---|---|---|---|---|---|
|  |  | Train | Test | Precision | Recall | Accuracy | IoU | F1 |
| Internal testing results | Worst Case | WC_1 | WC_2 | 0.94±0.03 | 0.92±0.08 | 0.94±0.04 | 0.87±0.08 | 0.93±0.05 |
|  |  |  | WC_3 | 0.92±0.05 | 0.94±0.05 | 0.94±0.03 | 0.87±0.07 | 0.93±0.04 |
|  |  | **WC_2** | WC_1 | 0.93±0.04 | 0.82±0.24 | 0.89±0.1 | 0.77±0.22 | **0.84±0.19** |
|  |  |  | WC_3 | 0.92±0.06 | 0.94±0.06 | 0.94±0.03 | 0.87±0.08 | 0.93±0.05 |
|  |  | WC_3 | WC_1 | 0.93±0.03 | 0.87±0.18 | 0.91±0.07 | 0.81±0.16 | 0.88±0.13 |
|  |  |  | WC_2 | 0.92±0.03 | 0.94±0.06 | 0.94±0.03 | 0.88±0.07 | 0.93±0.04 |
|  | Average Case | **AC_1** | AC_2 | 0.93±0.03 | 0.95±0.05 | 0.95±0.02 | 0.88±0.05 | **0.94±0.03** |
|  |  |  | AC_3 | 0.92±0.05 | 0.90±0.14 | 0.92±0.05 | 0.83±0.13 | 0.90±0.10 |
|  |  | AC_2 | AC_1 | 0.93±0.03 | 0.96±0.03 | 0.95±0.02 | 0.90±0.04 | 0.94±0.03 |
|  |  |  | AC_3 | 0.92±0.05 | 0.93±0.09 | 0.93±0.04 | 0.86±0.10 | 0.92±0.06 |
|  |  | AC_3 | AC_1 | 0.93±0.03 | 0.96±0.03 | 0.95±0.02 | 0.90±0.04 | 0.94±0.03 |
|  |  |  | AC_2 | 0.93±0.03 | 0.95±0.03 | 0.95±0.02 | 0.89±0.04 | 0.94±0.02 |
|  | Best Case | BC_1 | BC_2 | 0.92±0.04 | 0.93±0.08 | 0.94±0.03 | 0.87±0.08 | 0.93±0.05 |
|  |  |  | BC_3 | 0.93±0.03 | 0.92±0.09 | 0.93±0.05 | 0.86±0.09 | 0.92±0.06 |
|  |  | BC_2 | BC_1 | 0.91±0.06 | 0.97±0.03 | 0.95±0.02 | 0.89±0.06 | 0.94±0.04 |
|  |  |  | BC_3 | 0.92±0.04 | 0.95±0.06 | 0.94±0.04 | 0.88±0.07 | 0.93±0.04 |
|  |  | BC_3 | BC_1 | 0.91±0.06 | 0.96±0.03 | 0.94±0.03 | 0.88±0.07 | 0.94±0.04 |
|  |  |  | BC_2 | 0.92±0.04 | 0.95±0.05 | 0.94±0.02 | 0.87±0.06 | 0.93±0.04 |
| External testing dataset | Worst Case | WC_1 |  | 0.94±0.02 | 0.94±0.07 | 0.94±0.05 | 0.89±0.07 | 0.94±0.04 |
|  |  | **WC_2** |  | 0.94±0.02 | 0.9±0.17 | 0.93±0.05 | 0.85±0.16 | **0.91±0.13** |
|  |  | WC_3 |  | 0.94±0.02 | 0.93±0.1 | 0.93±0.05 | 0.87±0.09 | 0.93±0.06 |
|  | Average Case | AC_1 |  | 0.94±0.03 | 0.93±0.10 | 0.94±0.04 | 0.87±0.09 | 0.93±0.06 |
|  |  | AC_2 |  | 0.93±0.02 | 0.95±0.05 | 0.95±0.02 | 0.89±0.05 | 0.94±0.03 |
|  |  | AC_3 |  | 0.94±0.03 | 0.95±0.04 | 0.95±0.03 | 0.89±0.04 | 0.94±0.02 |
|  | Best Case | BC_1 |  | 0.94±0.02 | 0.94±0.07 | 0.94±0.04 | 0.88±0.06 | 0.94±0.04 |
|  |  | **BC_2** |  | 0.93±0.03 | 0.97±0.02 | 0.95±0.02 | 0.90±0.04 | **0.95±0.02** |
|  |  | BC_3 |  | 0.92±0.03 | 0.96±0.03 | 0.94±0.02 | 0.89±0.04 | 0.94±0.02 |

| Colon adenocarcinoma classification | | | | | | | | |
|---|---|---|---|---|---|---|---|---|
| | | Train | Test | Precision | Recall | Accuracy | IoU | F1 |
| Internal testing results | Worst Case | WC_1 | WC_2 | 0.92±0.19 | 0.85±0.24 | 0.83±0.22 | 0.81±0.24 | 0.87±0.22 |
| | | | WC_3 | 0.68±0.33 | 0.33±0.34 | 0.57±0.24 | 0.29±0.29 | 0.37±0.33 |
| | | WC_2 | WC_1 | 0.76±0.27 | 0.85±0.22 | 0.69±0.25 | 0.64±0.28 | 0.75±0.23 |
| | | | WC_3 | 0.53±0.30 | 0.68±0.36 | 0.52±0.24 | 0.43±0.30 | 0.54±0.31 |
| | | **WC_3** | WC_1 | 0.78±0.32 | 0.62±0.31 | 0.65±0.29 | 0.55±0.30 | 0.66±0.30 |
| | | | WC_2 | 0.28±0.44 | 0.02±0.07 | 0.12±0.15 | 0.02±0.07 | **0.03±0.10** |
| | Average Case | AC_1 | AC_2 | 0.89±0.22 | 0.79±0.29 | 0.82±0.20 | 0.73±0.29 | 0.80±0.27 |
| | | | AC_3 | 0.92±0.15 | 0.82±0.24 | 0.83±0.18 | 0.77±0.24 | 0.84±0.21 |
| | | AC_2 | AC_1 | 0.85±0.17 | 0.89±0.20 | 0.85±0.17 | 0.77±0.22 | 0.85±0.18 |
| | | | AC_3 | 0.89±0.17 | 0.88±0.18 | 0.85±0.16 | 0.80±0.21 | 0.87±0.17 |
| | | AC_3 | AC_1 | 0.86±0.18 | 0.87±0.22 | 0.85±0.17 | 0.78±0.23 | 0.85±0.20 |
| | | | AC_2 | 0.89±0.17 | 0.85±0.22 | 0.84±0.17 | 0.77±0.23 | 0.84±0.19 |
| | Best Case | **BC_1** | BC_2 | 0.91±0.17 | 0.90±0.18 | 0.89±0.14 | 0.83±0.21 | **0.89±0.18** |
| | | | BC_3 | 0.91±0.14 | 0.88±0.17 | 0.86±0.15 | 0.81±0.20 | 0.88±0.15 |
| | | BC_2 | BC_1 | 0.88±0.17 | 0.84±0.26 | 0.82±0.18 | 0.75±0.26 | 0.82±0.23 |
| | | | BC_3 | 0.90±0.16 | 0.88±0.18 | 0.86±0.16 | 0.80±0.21 | 0.87±0.16 |
| | | BC_3 | BC_1 | 0.82±0.21 | 0.88±0.23 | 0.84±0.16 | 0.76±0.25 | 0.84±0.21 |
| | | | BC_2 | 0.85±0.20 | 0.88±0.23 | 0.84±0.19 | 0.78±0.26 | 0.85±0.22 |
| External testing dataset | Worst Case | WC_1 | | 0.89±0.13 | 0.86±0.18 | 0.84±0.13 | 0.77±0.19 | 0.86±0.15 |
| | | **WC_2** | | 0.45±0.38 | 0.26±0.35 | 0.31±0.22 | 0.19±0.25 | **0.26±0.31** |
| | | WC_3 | | 0.89±0.13 | 0.79±0.27 | 0.79±0.22 | 0.71±0.26 | 0.80±0.23 |
| | Average Case | AC_1 | | 0.89±0.14 | 0.75±0.29 | 0.78±0.23 | 0.69±0.28 | 0.77±0.26 |
| | | AC_2 | | 0.90±0.12 | 0.80±0.23 | 0.81±0.21 | 0.73±0.23 | 0.82±0.19 |
| | | AC_3 | | 0.92±0.11 | 0.82±0.22 | 0.84±0.17 | 0.76±0.21 | 0.84±0.16 |
| | Best Case | **BC_1** | | 0.91±0.12 | 0.92±0.09 | 0.89±0.11 | 0.85±0.14 | **0.91±0.09** |
| | | BC_2 | | 0.9±0.13 | 0.87±0.19 | 0.84±0.15 | 0.78±0.18 | 0.86±0.13 |
| | | BC_3 | | 0.86±0.14 | 0.85±0.21 | 0.83±0.15 | 0.76±0.22 | 0.84±0.17 |

| Rectal cancer segmentation | | | | | | | | |
|---|---|---|---|---|---|---|---|---|
| | | Train | Test | Precision | Recall | Accuracy | IoU | F1 |
| Internal testing results | Worst Case | WC_1 | WC_2 | 0.70±0.24 | 0.67±0.27 | 0.97±0.06 | 0.47±0.21 | 0.61±0.22 |
| | | | WC_3 | 0.69±0.26 | 0.68±0.30 | 0.96±0.07 | 0.50±0.24 | 0.63±0.26 |
| | | WC_2 | WC_1 | 0.74±0.26 | 0.56±0.27 | 0.97±0.03 | 0.45±0.22 | 0.59±0.24 |
| | | | WC_3 | 0.66±0.29 | 0.60±0.30 | 0.96±0.04 | 0.45±0.24 | 0.58±0.26 |
| | | WC_3 | WC_1 | 0.70±0.25 | 0.69±0.26 | 0.97±0.02 | 0.52±0.23 | 0.65±0.23 |
| | | | WC_2 | 0.65±0.27 | 0.77±0.25 | 0.97±0.04 | 0.51±0.22 | 0.64±0.22 |
| | Average Case | AC_1 | AC_2 | 0.72±0.25 | 0.68±0.25 | 0.98±0.02 | 0.50±0.21 | 0.64±0.21 |
| | | | AC_3 | 0.73±0.25 | 0.64±0.27 | 0.96±0.05 | 0.51±0.22 | 0.64±0.23 |
| | | AC_2 | AC_1 | 0.62±0.27 | 0.75±0.29 | 0.96±0.03 | 0.48±0.23 | 0.61±0.24 |
| | | | AC_3 | 0.70±0.26 | 0.70±0.30 | 0.96±0.07 | 0.50±0.23 | 0.63±0.23 |
| | | **AC_3** | AC_1 | 0.43±0.26 | 0.82±0.24 | 0.92±0.07 | 0.37±0.22 | **0.50±0.25** |
| | | | AC_2 | 0.47±0.27 | 0.83±0.22 | 0.94±0.06 | 0.40±0.23 | 0.54±0.24 |
| | Best Case | BC_1 | BC_2 | 0.69±0.26 | 0.72±0.28 | 0.97±0.06 | 0.52±0.23 | 0.64±0.24 |
| | | | BC_3 | 0.74±0.23 | 0.68±0.29 | 0.96±0.07 | 0.51±0.23 | 0.64±0.23 |
| | | **BC_2** | BC_1 | 0.71±0.23 | 0.73±0.25 | 0.98±0.02 | 0.54±0.21 | **0.68±0.21** |
| | | | BC_3 | 0.74±0.22 | 0.68±0.28 | 0.96±0.07 | 0.51±0.23 | 0.64±0.22 |
| | | BC_3 | BC_1 | 0.57±0.25 | 0.86±0.19 | 0.96±0.02 | 0.50±0.22 | 0.63±0.22 |
| | | | BC_2 | 0.56±0.27 | 0.84±0.20 | 0.96±0.04 | 0.47±0.22 | 0.61±0.22 |
| External testing dataset | Worst Case | WC_1 | | 0.66±0.26 | 0.71±0.28 | 0.98±0.01 | 0.52±0.24 | 0.64±0.24 |
| | | **WC_2** | | 0.58±0.36 | 0.60±0.37 | 0.98±0.01 | 0.43±0.29 | **0.54±0.33** |
| | | WC_3 | | 0.62±0.24 | 0.86±0.18 | 0.98±0.02 | 0.55±0.21 | 0.69±0.20 |
| | Average Case | AC_1 | | 0.66±0.24 | 0.81±0.16 | 0.98±0.01 | 0.55±0.20 | 0.69±0.19 |
| | | AC_2 | | 0.65±0.27 | 0.79±0.24 | 0.98±0.01 | 0.53±0.23 | 0.66±0.23 |
| | | AC_3 | | 0.45±0.26 | 0.88±0.19 | 0.94±0.07 | 0.42±0.23 | 0.55±0.25 |
| | Best Case | BC_1 | | 0.65±0.25 | 0.81±0.22 | 0.98±0.01 | 0.56±0.22 | 0.69±0.22 |
| | | **BC_2** | | 0.68±0.22 | 0.82±0.18 | 0.98±0.01 | 0.58±0.19 | **0.71±0.17** |
| | | BC_3 | | 0.56±0.24 | 0.90±0.13 | 0.97±0.02 | 0.52±0.21 | 0.65±0.21 |

Extended Data Table 4 Detailed performance measures for each experiment for all three use cases. For the F1 score, the best performance is highlighted in red, and the worst performance is highlighted in blue.

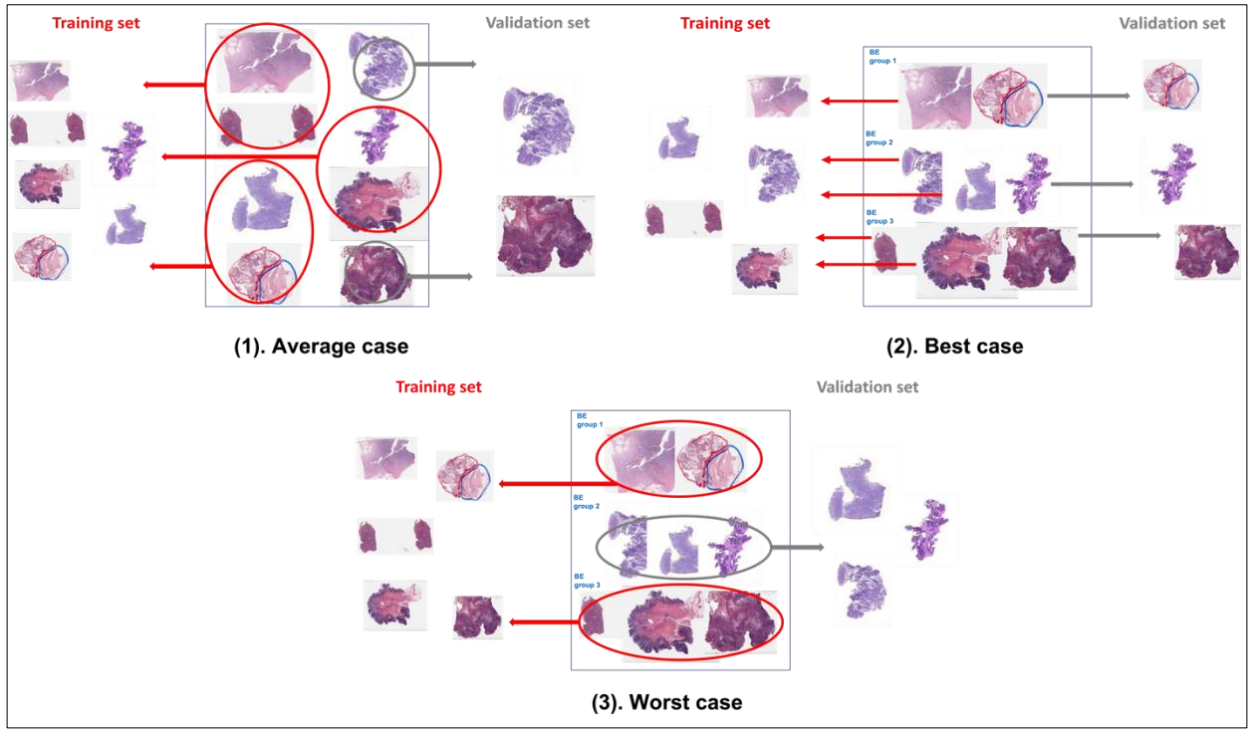

Extended Data Figure 1 Three data partitioning strategies. **(1) Average case**: Patients are randomly split into training and testing sets without considering BEs, which can cause possible sub-optimal situations. **(2) Worst case**: patients with similar BE metrics are exclusively assigned to the training or testing dataset, resulting in slides in the training set looking highly dissimilar to those in the testing set. **(3) Best case**: where detected BE-groups are systematically divided between training/testing sets. This process, enabled via CF, ensures the diversity of the training dataset, and thus, improves the robustness of the machine model.

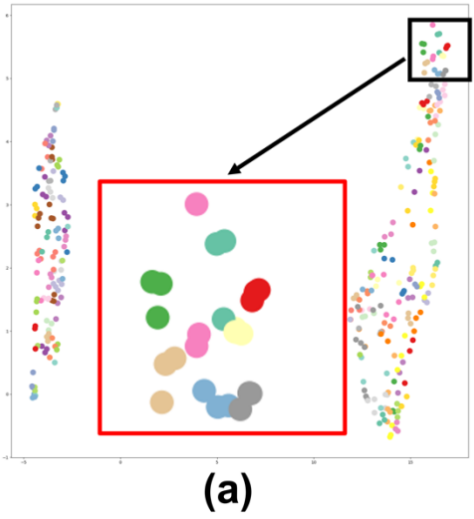 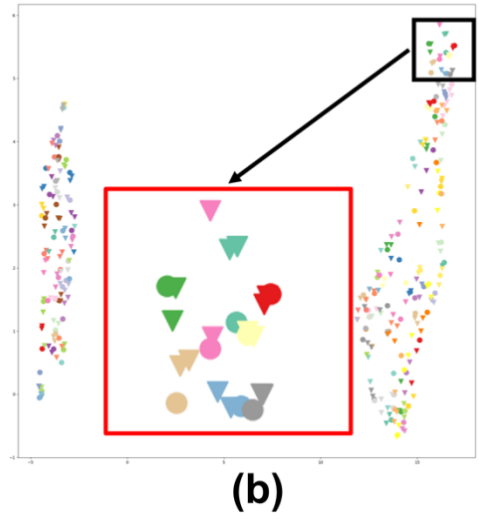

Extended Data Figure 2 Detail of UMAP plots generated via CohortFinder for the colon adenocarcinoma classification use case **(a)** Quality measures embedded in a 2-dimensional plot using U-MAP. **(b)** Dots replaced with "v (triangle-down)" and "o (circle)", where "v" indicates a slide CF for inclusion in the training set and "o" indicating a slide to be placed in the testing set.

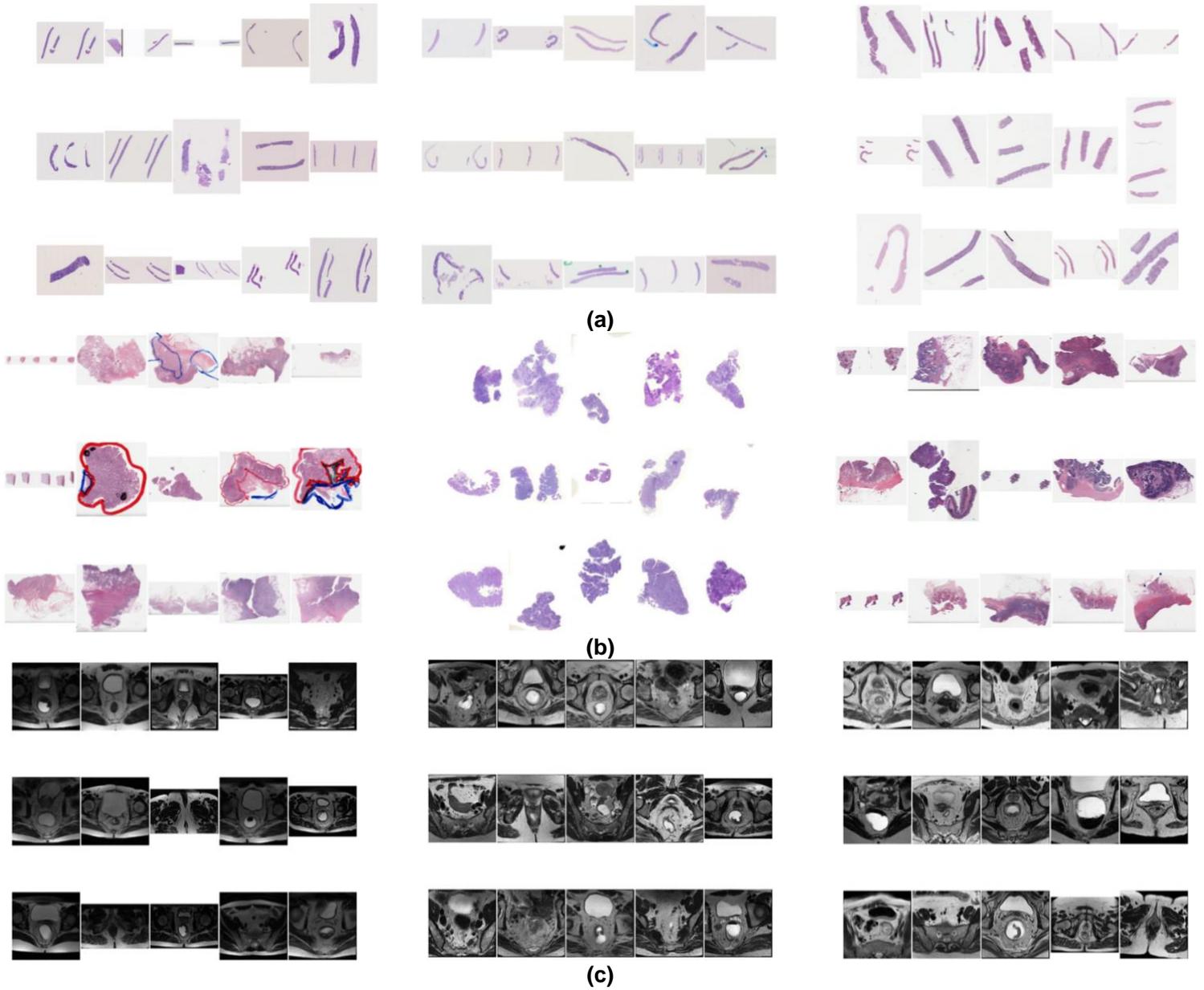

Extended Data Figure 3 Three contact sheets produced by CohortFinder for **(a)** Tubule segmentation on WSIs **(b)** Colon adenocarcinoma classification on WSIs **(c)** Rectal cancer segmentation on MRIs. Each row shows 3 detected BE-groups with notable (i) intra group homogeneity and (ii) inter-group heterogeneity, providing a visual confirmation of successful BE-group detection.

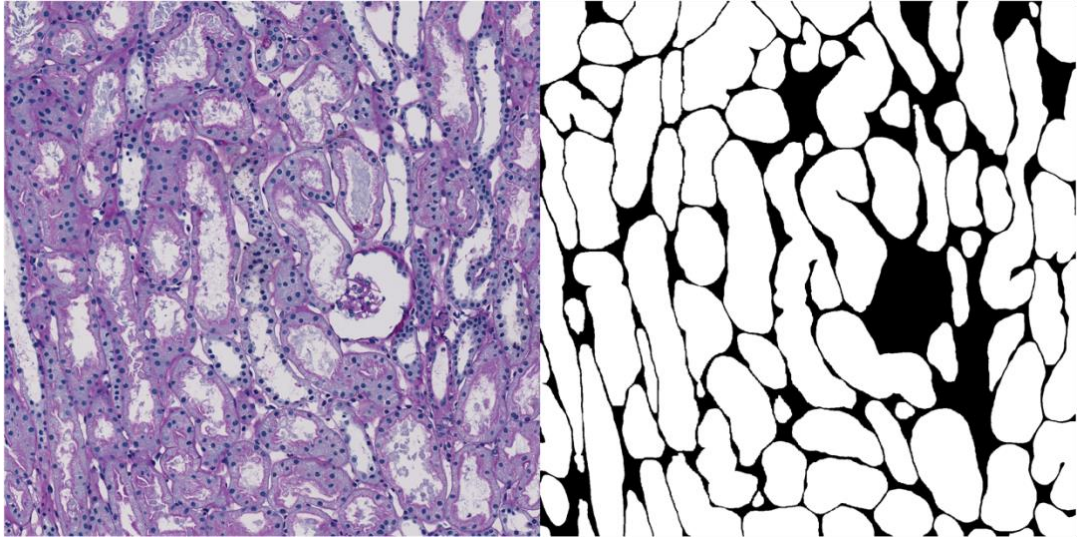

Extended Data Figure 4 Ground truth example for the tubule segmentation use case, the left one is the cropped ROI, and the right one is the expert annotation.

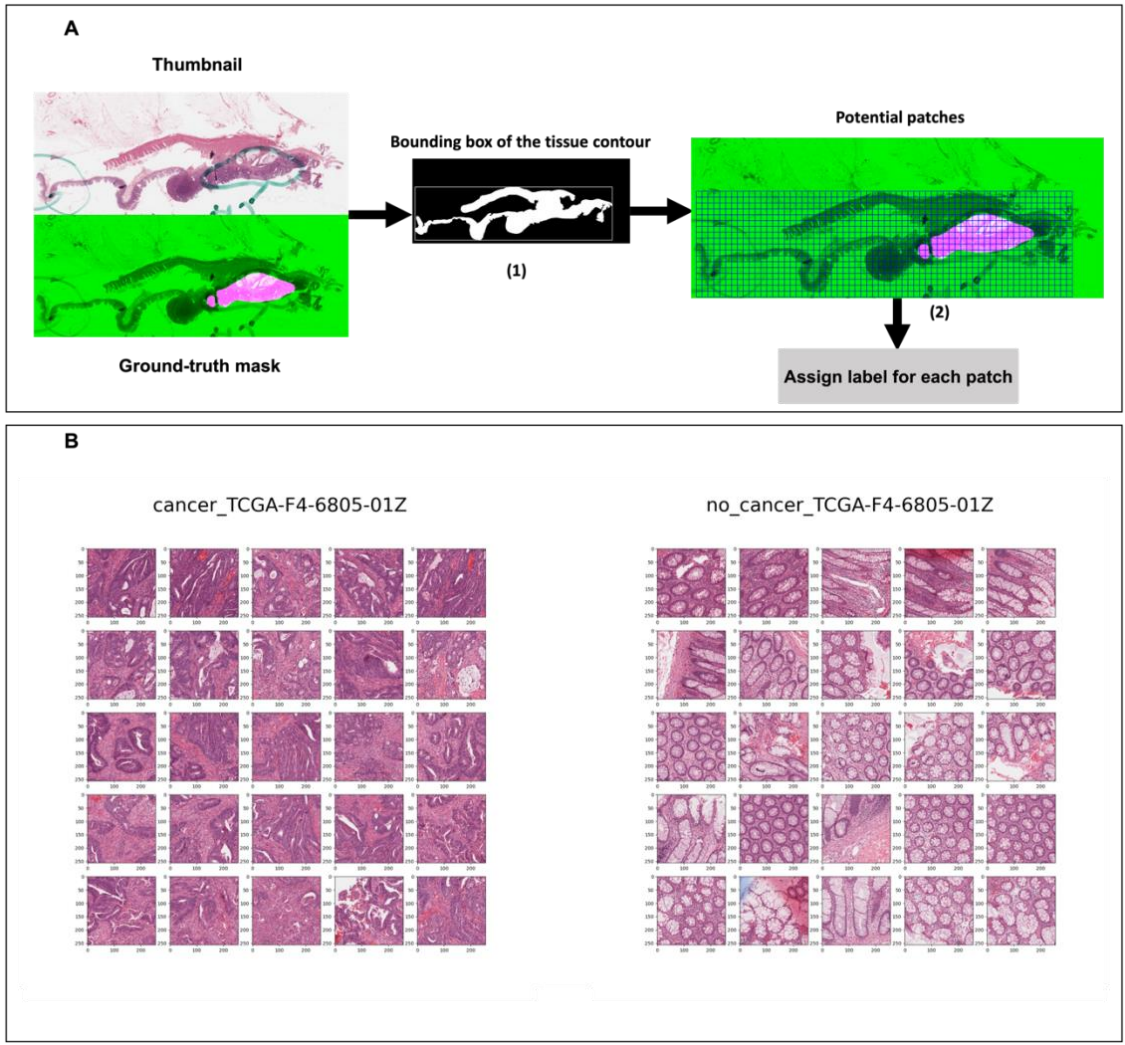

Extended Data Figure 5 Patch sampling process and result for the colon cancer classification use case. **A**: Patch sampling process, **B**: Patch examples for cancer vs no cancer from patient: TCGA-F4-6805-01Z.

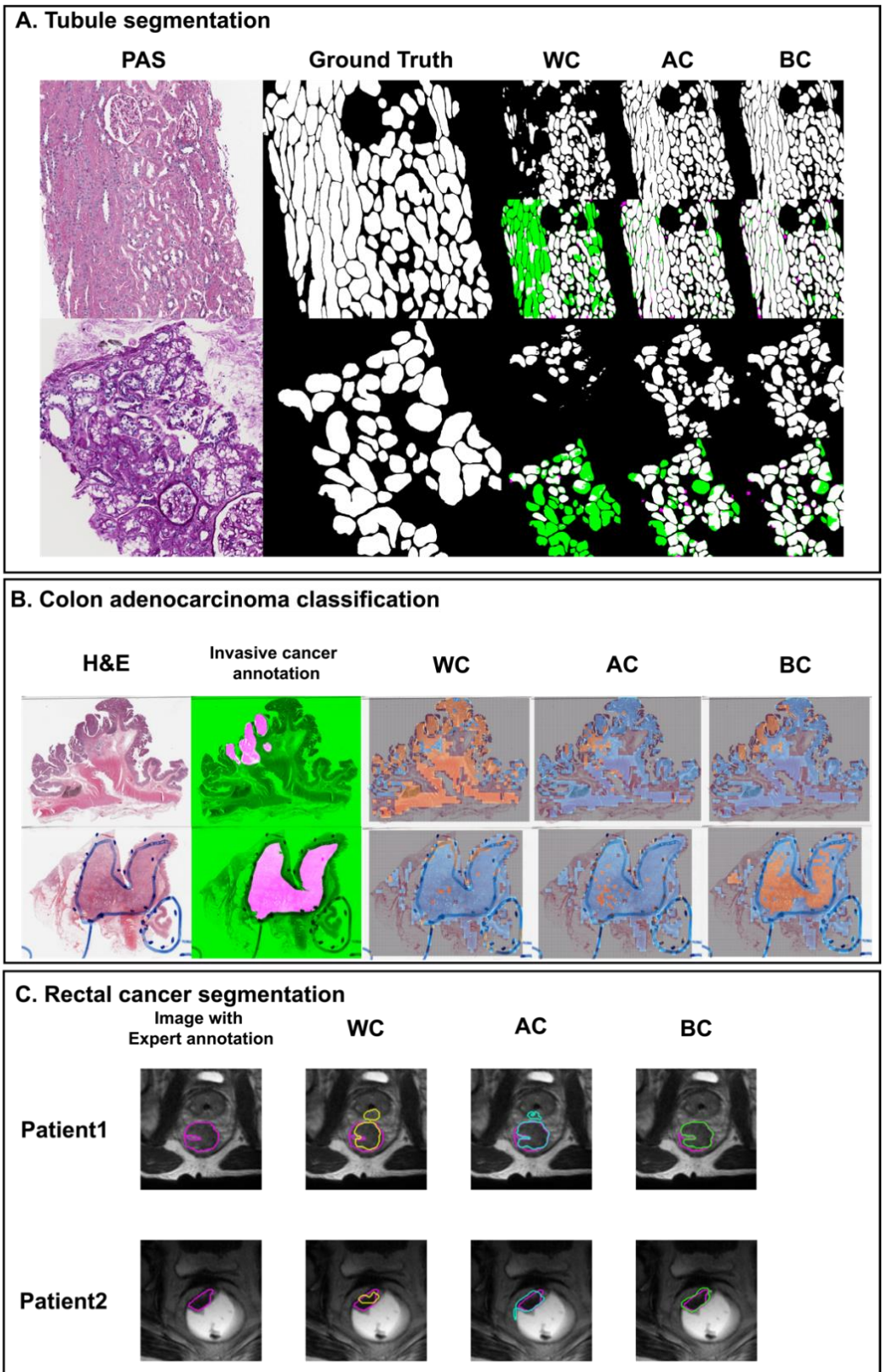

Extended Data Figure 6 Qualitative comparison of the three use cases in the external set. **A) For tubule segmentation task**: The first column is a 3000x3000 PAS-stained ROI cropped at 40x, the second column is the tubule segmentation ground truth (GT). The remaining images are the results of BC, AC and WC. In each scenario, the top row is the DL model results, while the bottom row corresponds to the overlay image of DL output images with the GT, where green parts represent the false negative (FN) area, and the red parts represents the false positive (FP) area. **B) For the colon cancer classification task**: The first column represents the thumbnails of the H&E stained WSI, the second column is the cancer

region annotation overlaid in pink. Remaining images are the heatmap results of different BC, AC, and WC, where the detected cancer area is highlighted as orange, the non-cancer area is highlighted as blue, while the gray area corresponds to non-informative (representing background or non-tissue). **C) For the rectal tumor segmentation task**: The first column corresponds to the patient MRI with the expert tumor annotation contoured in pink. The remaining images are rectal tumor segmentations via WC (yellow), AC (cyan), and BC (green) models compared to expert annotations (pink).

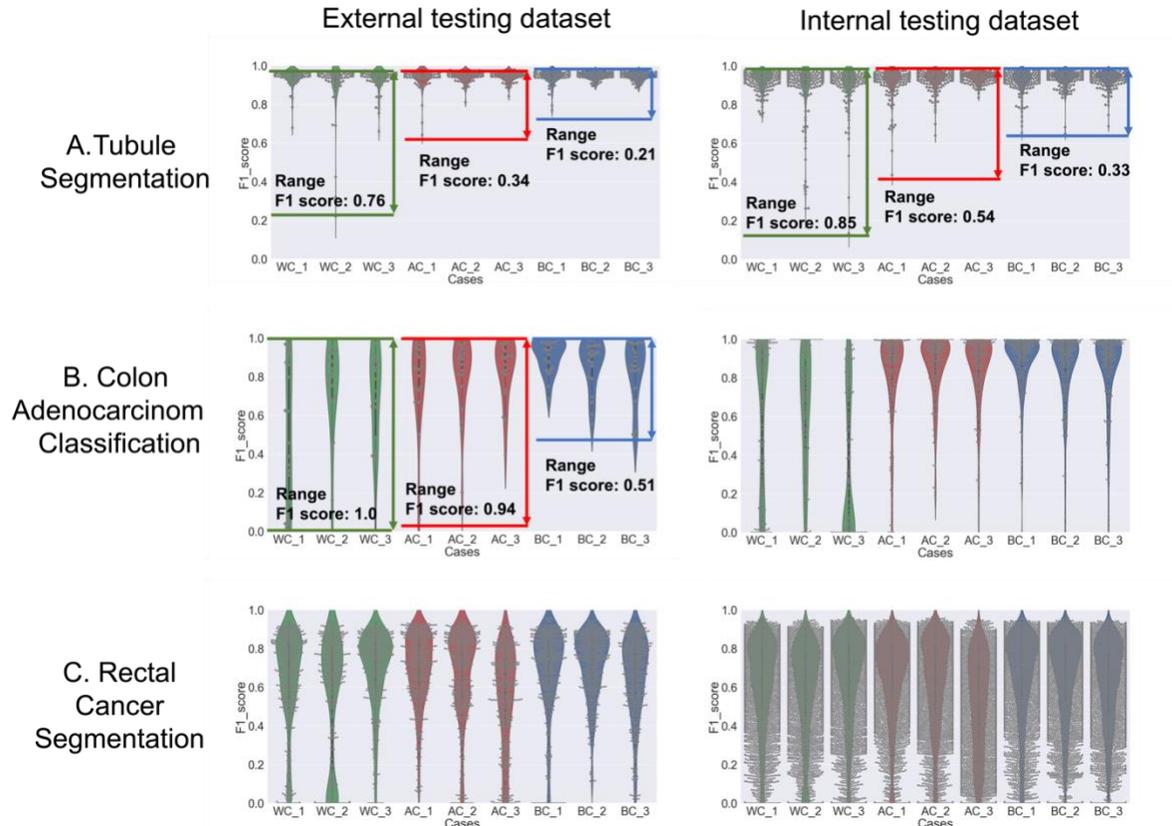

Extended Data Figure 7 Violin plots of the F1-score evaluation measure distribution for both external and internal testing dataset and for each use case. Each dot represents the F1-score value for **(a)** tubule segmentation at ROI level, **(b)** colon adenocarcinoma classification at WSI level and **(c)** rectal cancer segmentation at slice level. In a majority of the comparisons, the BC distribution appears more compact than WC and AC distributions.